\documentclass[10pt,final,conference,letterpaper,romanappendices,twocolumn]{IEEEtran}
\IEEEoverridecommandlockouts
\usepackage{enumerate}
\usepackage{graphicx}
\usepackage{epsf}
\usepackage{subfigure}
\usepackage{amsmath}
\usepackage{amssymb}
\usepackage{array}
\usepackage{setspace}
\usepackage[amsmath,thmmarks,amsthm]{ntheorem}
\usepackage[ruled,lined]{algorithm2e}
\usepackage{algorithmic}
\usepackage{amsmath, bm}

\usepackage{xcolor}
\usepackage{amsfonts}
\usepackage{dsfont}
\usepackage{xfrac}
\usepackage{breqn}
\usepackage{bbm}
\usepackage{cite}

\usepackage{graphicx}
\graphicspath{{Fig/}}

\newcommand{\defn}{\triangleq}

\theoremseparator{.}

\begin{document}
	
\title{Federated Deep Reinforcement Learning for THz-Beam Search with Limited CSI}
	

\author{\IEEEauthorblockN{Po-Chun Hsu, Li-Hsiang Shen, Chun-Hung Liu$^{*}$, and Kai-Ten Feng}\\
\IEEEauthorblockA{Department of Electrical and Computer Engineering, National Yang Ming Chiao Tung University, Hsinchu, Taiwan\\
Department of Electrical and Computer Engineering, Mississippi State University, MS, USA$^{*}$}
e-mail: \{g309513013.c, ktfeng, gp3xu4vu6.cm04g\}@nycu.edu.tw and chliu@ece.msstate.edu$^{*}$}

\maketitle

\begin{abstract}
Terahertz (THz) communication with ultra-wide available spectrum is a promising technique that can achieve the stringent requirement of high data rate in the next-generation wireless networks, yet its severe propagation attenuation significantly hinders its implementation in practice. Finding beam directions for a large-scale antenna array to effectively overcome severe propagation attenuation of THz signals is a pressing need. This paper proposes a novel approach of federated deep reinforcement learning (FDRL) to swiftly perform THz-beam search for multiple base stations (BSs) coordinated by an edge server in a cellular network. All the BSs conduct deep deterministic policy gradient (DDPG)-based DRL to obtain THz beamforming policy with limited channel state information (CSI). They update their DDPG models with hidden information in order to mitigate inter-cell interference. We demonstrate that the cell network can achieve higher throughput as more THz CSI and hidden neurons of DDPG are adopted. We also show that FDRL with partial model update is able to nearly achieve the same performance of FDRL with full model update, which indicates an effective means to reduce communication load between the edge server and the BSs by partial model uploading. Moreover, the proposed FDRL outperforms conventional non-learning-based and existing non-FDRL benchmark optimization methods. 
\end{abstract}

\footnotetext[1]
{The work of P.-C. Hsu, L.-H. Shen, and K.-T. Feng was supported in part funded by Ministry of Science and Technology (MoST) Grants 110-2221-E-A49-041-MY3, 111-2218-E-A49-024, STEM Project, the National Defense Science and Technology Academic Collaborative Research Project in 2022, and Higher Education Sprout Project of the National Yang Ming Chiao Tung University and Ministry of Education (MoE), Taiwan. The work of C.-H. Liu was supported in part by the U.S. National Science Foundation (NSF) under Award CNS-2006453.}

\section{Introduction}

For the purpose of meeting the increasing ultra-high data requirement, such as virtual reality (VR), augmented reality (AR) and hologram technologies in the future sixth generation (6G) communication network, the prospect of terahertz (THz) communication is considered to be able to provide higher frequency bands ranging from 0.1 to 10 THz. However, compared to conventional millimeter wave (mmWave) at gigahertz (GHz) bands, the most different and important challenges for THz are severe power attenuation, blockages and additional molecular absorption that leads to a much shorter propagation distance. Therefore, beamforming techniques are utilized to enhance the transmit direction towards the desired receiving user equipment (UE) rather than omni-directional ones which do not require a large-scale antenna array deployed to obtain high beam gain and spatial diversity. Although there exist potentially rich channel paths in a multi-antenna system, only a single line-of-sight (LoS) link can be utilized in the THz band in most cases. The related beamforming designs in \cite{NBCZCW21}, \cite{HCYZAG20} were proposed for THz systems. However, when associating with enormous UEs in different cells, serious interfered beamforming issues would probably exist in short transmission distances of a THz network, and thereby it would be necessary to appropriately coordinate multiple BSs in the THz network so as to improve the overall transmission performance of the THz network. In addition, references in \cite{TPVTD18,TG12,SFYW17} employed full CSI of beamforming that requires highly complicate and accurate channel estimation, which is fairly difficult to be implemented for dynamic large-scale antenna arrays. Furthermore, estimating the full CSI of a large-scale antenna array is always a difficult task because wireless channels are prone to be affected by dynamic environmental variances. Accordingly, traditional optimization methods are hardly to perfectly tackle the beam search problem in wireless network deployed with large-scale antenna arrays of THz.

Recently, deep learning techniques are widely applied in the different fields in wireless communication systems. As a prospect, deep reinforcement learning (DRL) enables the agent, which may be BS or UE to adjust its wireless state and action, i.e., policy output according to the changed environment. Different from model-free reinforcement learning, the deep Q network (DQN) architecture is implemented via deep neural network (DNN) to decide the Q-value instead of the Q-table,  which benefits problems with non-countable or near-continuous variables with infinite solution sets. However, when the action space is high-dimensional and continuous, it is very inefficient to apply a regular DQN to obtain the decision space. Deep deterministic policy gradient (DDPG) using a two-layered DNN as actor-critic (AC) network is known to deal with this problem. Federated edge learning (FEL) has been considerably studied for improving the training progress via learning model exchange with less information uploaded to edge server \cite{YSLY20}. The main concept of FEL is to integrate local training models from different (mobile) clients in order to acquire a more complete global model, which can include certain hidden information in different clients. References in~\cite{MXXJ21,LKHHYH21,ZSTH21} studied how to improve the learning speed when minimizing computing delay and energy consumption through FEL. In  \cite{LWYBL20}, how to enhance the privacy of every participating client by using FEL. Although these aforementioned prior works advanced the study of FEL over wireless communication, they overlooked two main practical issues that were not considered in their system models, i.e., co-channel interference and limited CSI for large-scale antenna array. Therefore, there exists a pressing need to design interference mitigation scheme in THz beamforming by combining FEL with deep learning techniques. 

The contributions of this paper are summarized as follows.
\begin{itemize}
\item We have conceived a federated deep reinforcement learning (FDRL) leveraging the benefits of both FEL and DRL architectures. FEL aims at model exchange from neural networks extracting hidden information of partially estimated CSI, which potentially alleviates interference from other BSs. While, AC-based DDPG is designed to search candidate THz beams to maximize the total throughput performance.
\item We characterize the performance in terms of computational complexity and network throughput. We can infer that higher network throughput can be achieved with more antennas, exchanged data, and more neurons of FDRL under a compromised computational complexity of deep learning. The proposed FDRL scheme outperforms the baseline using pure deep Q-learning and conventional non-deep learning based beamforming methods.
\end{itemize}

The rest of the paper is organized as follows: Section \uppercase\expandafter{\romannumeral2} describes the system model and formulates the THz beamforming problem. Section \uppercase\expandafter{\romannumeral3} elaborates on our proposed FDRL algorithm for coordinating THz beamforming under a multi-BS network. Section \uppercase\expandafter{\romannumeral4} shows simulation results, whlist conclusions are drawn in Section \uppercase\expandafter{\romannumeral5}.
	
\section{System Model} \label{SM}
	
In this paper, we consider a cellular network in which each UE is equipped with a single antenna and there are $K$ base stations (BSs) operating in the THz (frequency) band, each of which equipped with $N$ antennas. In the downlink, each BS is assumed to adopt different resource blocks to serve different UEs in its cell. Namely, no UEs in the same cell share the same resource blocks and thereby there is no intra-cell interference in the network. We also assume that the frequency reuse factor is one in the network so that the $K$ BSs interferes each other in the downlink and UE $k$ thus receives interference from the other $K-1$ BSs when it is served by BS $k$. All the BSs are connected with high-speed optical links to an edge server where edge computing can be conducted. A schematic diagram of the cellular network with edge computing considered in this paper is shown in Fig. 
1. In the following, we will first introduce the channel model in the THz band and then specify the signal model transmitted over a THz channel.

\subsection{THz Channel Model}

Due to THz signals' nature of extremely high frequency, transmitting them significantly suffers from two serious environmental impairments, i.e., severe attenuation and molecular absorption~\cite{LCLGY15,PSKT13,HCBAOAIF15,GXDLZY17}. As such, THz signals undergo much higher path loss than mmWave as well as UHF signals. For a THz channel with frequency $f$, its channel response for transmitting a signal over distance $d$, denoted by complex vector $\mathbf{h}\in\mathbb{C}^{N}$, can be modeled as

\begin{align}\label{Eqn:ChannelModel}
	\mathbf{h}=G \left[1+\sum_{l=1}^{L} \Lambda_l(f)\right]a_{L}(f,d) \mathbf{a}_t(\theta_{t}),
\end{align}
where $G$ is called the integrated antenna gain consisting of the transmitted and received antenna gains of the antenna array, $L$ denotes the number of non-line-of-sight (NLoS) paths, $\Lambda_l(f)$ is a frequency-dependent constant consisting of the reflection factor and roughness coefficient of NLoS paths affected by the reflective interfaces and material impedance. Moreover, $a_{L}(f,d)$ is defined as 
\begin{align*}
a_{L}(f,d)\defn \frac{c}{4 \pi f d} e^{-\frac{1}{2} \rho(f) d}
\end{align*}  
in which $c$ is the speed of light and $\rho(f)$ is the medium absorption factor of frequency $f$. Furthermore, we consider a uniform linear array at each BS so that $\mathbf{a}_{t}(\theta_t)$ can be defined as
\begin{align}\label{antenna}
\hspace{-7pt}\mathbf{a}_{t}(\theta_t)\defn\frac{1}{N}\left[1,e^{j\frac{2 \pi}{\lambda}d_a \sin(\theta_t)},\cdots,e^{j\frac{2 \pi}{\lambda}d_a (N-1)\sin(\theta_t)}\right]^{T},
\end{align}
where $\theta_t\in[-\frac{\pi}{2},\frac{\pi}{2}]$ is the angle of departure, $d_a$ is the distance between two antennas, and $T$ denotes the transpose operation of a vector. Note that $\mathbf{h}$ in~\eqref{Eqn:ChannelModel} consists of LoS and NLoS components, that is, $Ga_L(f,d)\mathbf{a}(t)(\theta_t)$ is the LoS component whereas the other term is the NLoS component. 

\begin{figure}\label{Fig.1}
	\centering
	\includegraphics[width=3.25in, height=1.85in]{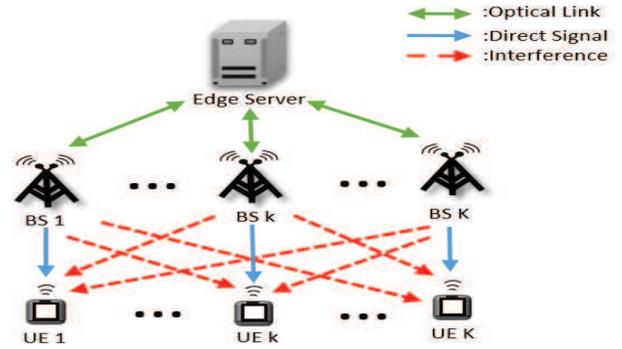}
	\vspace{-0.1in}
	\caption{A schematic diagram of the cellular network with edge computing considered in this paper. In the network, there are $K$ base stations, each of them equipped with $N$ transmit antennas. A downlink communication scenario is considered and BS $k$ serves UE $k$ equipped with a single antenna in the THz frequency band. Each of the BSs is connected to an edge server through a high-speed optical link.} 
	\vspace{-0.22in}
\end{figure}

\subsection{THz Signal Model}

Let $\mathbf{w}_k \in \mathbb{C}^{N}$ be the beamforming vector for BS $k$ and $x_k \in \mathbb{C}$ be the signal with unit power transmitted to the $k$th UE by BS $k$. Since there are $K$ BSs in the network, we consider the worst scenario that all the BSs interferes each other when they serve their UE. As a result, the signal received by UE $k$ can be specifically written as
\begin{align}\label{recieved signal}
y_k= \underbrace{\sqrt{P}\, \mathbf{h}_{kk}^{\mathcal{H}}\mathbf{w}_kx_k}_{\text{desired signal}} +\underbrace{\sum_{j=1,j\neq i}^{K} \sqrt{P}\, \mathbf{h}_{jk}^{\mathcal{H}}\mathbf{w}_{j}x_{j}} _{\text{interference signal}} + \underbrace{n_k}_{\text{noise}},
\end{align}	
where $k\in\{1,\ldots,K\}$, $P$ is the transmit power of each BS, and superscript $\mathcal{H}$ stands for the Hermitian operation of a complex vector, $n_k\in\mathbb{C}$ denotes the Gaussian noise, and $\mathbf{h}_{kk} \in \mathbb{C}^{N}$ and $\mathbf{h}_{jk} \in \mathbb{C}^{N}$ are the channel vectors from BS $k$ to UE $k$ and from (interfering) BS $j$ to UE $k$, respectively. Note that $\mathbf{h}_{kk}$ and $\mathbf{h}_{jk}$ adopt the channel model defined in~\eqref{Eqn:ChannelModel}. As such, the signal-to-noise-plus-interference ratio (SINR) received at the $k$th UE can be defined as
\begin{align}\label{Eqn:SINR}
	\Gamma_k= \frac{P\,|\mathbf{h}_{kk}^{\mathcal{H}}\mathbf{w}_k|^2}{ \sum_{j=1,j\neq i}^{K} P\,|\mathbf{h}_{jk}^{\mathcal{H}}\mathbf{w}_{j}|^2 + \sigma_{n}^2},
\end{align}	 
where $|\cdot|$ represents the operator of absolute value and $\sigma_{n}^2$ is the power of the Gaussian noise $n_k$ for all $k\in\{1,\ldots,K\}$. According to~\eqref{Eqn:SINR}, the downlink achievable rate (spectral efficiency) of BS $k$ can be written as
\begin{align}
C_k=\log_{2}\left({1+\Gamma_k}\right),\quad\text{(bits/sec/Hz)}
\end{align}	 
for all $k\in\{1,\ldots,K\}$. In the following, we will use $C_k$ to formulate an optimization problem of beam search that is able to maximize the sum rate of all the BSs in the scenario that only limited CSI is available at each BS. 

\begin{figure*}[!t]
	\centering
	\includegraphics[width=6.5in, height=1.8in]{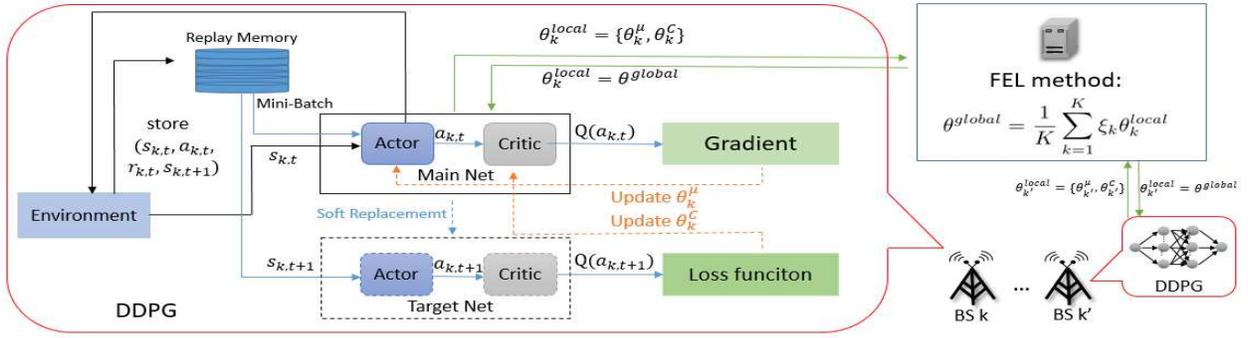}
	\caption{The DRL-based DDPG algorithm. The main/target networks contain their actor-critic sub-networks established by DNN.} \label{Fig.2}
	\vspace{-0.2in}
\end{figure*}

\subsection{Sum-Rate Optimization with Limited CSI} 

In this work, we aim to maximize the sum rate of the network by optimizing each beamforming vector $\mathbf{w}_k$ with limited CSI $g(\mathbf{h}_{kk})=\mathbf{h}_{kk}^{lim}\in \mathbb{C}^{N}$ that is estimated at each BS $k$. The limited CSI means that only part of the total CSI at each BS can be effectively estimated due to the large size of each antenna array, and thereby the achievable SINR $\Gamma_k$ in~\eqref{Eqn:SINR} reduces to the limited SINR $\Gamma_k^{lim}$, which equals $\Gamma_k$ that adopts $\mathbf{h}^{\lim}_{kk}$ in place of $\mathbf{h}_{kk}$. As a result, we can formulate a optimization problem of the (mean) network throughput (sum rate) with limited CSI as follows:
\begin{align} \label{capacity}
&\mathop{\max}\limits_{ \mathbf{w}_k } \ \sum_{k=1}^{K} C_k^{lim}\\
	&\text{s.t. }
	\mathrm{ tr}\left(\mathbf{w}_k\mathbf{w}_k^{\mathcal{H}}\right) \le 1,\,\,{g(\mathbf{h}_{kk}) = \mathbf{h}_{kk}^{lim}}, \,\, k\in\{1,\ldots,K\}, \nonumber
\end{align} 
where $C_k^{lim}\defn \log_{2}({1+\Gamma^{lim}_k})$ and $\mathrm{tr}(\cdot)$ denotes the trace operator of a matrix. However, the optimization problem in~$\eqref{capacity}$ cannot be readily solved via conventional optimization methods with respect to the digital beamforming and partially-attainable CSI. Furthermore, highly computational complexity of global optimum and huge overhead of CSI exchange caused by many antenna arrays in terahertz networks make the conventional methods fairly difficult to analyze the sophisticated and unpredictable communication network. As a consequence, we design a deep-learning-based scheme by jointly leveraging DRL and federated learning architectures, which helps to reduce the amount of CSI exchange and tries to relieve the impact of interference, to resolve the complex optimization problem.

\section{Federated Deep Reinforcement Learning (FDRL) for THz beam Search} \label{BP}
Since the optimization problem of the network throughput with limited CSI in~\eqref{capacity} is not analytically tractable, we propose an FEL-based DDPG learning scheme that can iteratively coordinate to attain an optimal policy of THz beamforming. We consider that each BS conducts a DDPG to obtain a policy of THz beamforming with limited CSI when all the BSs are controlled by a single FEL server to exchange training model with hidden information so as to mitigate interference. 

\subsection{DRL-based DDPG Network}
We consider that the DRL framework contains a state set $\mathcal{S}$, an action set $\mathcal{A}$, a reward set $\mathcal{R}$, and an agent (BS or UE) that applies a certain action to obtain the corresponding reward while updating the current status. The action will be reinforced iteratively in order to receive better rewards in a varying environment. 
On account of the large state-action set existing in our THz beamforming problem, using conventional DRL approaches to tackle the problem is certainly inappropriate because it definitely suffers from huge storage of table-mapping and slow convergence. Moreover, since a THz beamforming vector is deemed to be continuous variables with substantially-high quantization levels, this thus motivates us to adopt DDPG to establish a two-layered actor-critic network to resolve the problem with continuous solutions. For the DRL-based DDPG in the THz network, we define the state, action and rewards as follows.
	
	1)\emph{ State set $\mathcal{S}$}: In THz, the state set collects the statuses of each BS under current THz channels, denoted by $\mathcal{S} = \left\lbrace{s_k | k=1,2,...,K}\right\rbrace$, which consists of the serving CSI $\mathbf{h}_{kk}$ linked to the $i$-th UE, and SINR $\Gamma_k$ fedback from the UE $i$. Note that $\mathbf{h}_{kk}$ may be partially attainable due to limited measured CSI under a large-scale THz antenna array. Therefore, state of each BS should be $s_k =\left\lbrace{ \mathbf{h}_{kk}^{lim}, \Gamma_k | k=1,2,...,K}\right\rbrace$.
	
	2)\emph{ Action set $\mathcal{A}$}: The action set represents the decision-making of THz beamfoming vector defined as $\mathcal{A}=\left\lbrace{ \mathbf{a}_k=\mathbf{w}_k| k=1,2,...,K }\right\rbrace$. Note that each BS will only determine its own action, i.e., $\mathbf{w}_k$ according to current input state and reward.
	
	3)\emph{ Reward set $\mathcal{R}$}: We define the overall reward as $\mathcal{R} = \left\lbrace{r_k | k=1,2,...,K }\right\rbrace$. Since we aim at maximizing the sum throughput in $\eqref{capacity}$, we consider the reward function as individual throughput of each BS, i.e., $r_k= C_k^{lim}$.

As shown in Fig. \ref{Fig.2}, the DDPG architecture contains the main and target networks that individually consist of actor and critic sub-networks wherein $\theta^{\mu}_k$ and $\theta^{Q}_k$ denote DNN-enabled actor/critic weights in the main network, and $\theta^{\mu'}_k$ and $\theta^{Q'}_k$ denote the  actor/critic weights in the target network. The main network determines the beamforming action of the $i$-th THz BS as $a_{k,t}=\mu(s_{k,t}|\theta^{\mu}_k)+N_{G}$, where $\mu(s_{k,t}|\theta^{\mu}_k)$ is the output layer of the DNN-based actor network. To perform the exploration of the environment, the deterministic policy will obtain the probabilistic action by adding the perturbation $N_{G}$ as Gaussian noise. On the other hand, the target network input is fed by the action of actor network, which provides the Q-value outcome $Q\left(s_{k,t},a_{k,t}|\theta^{Q}_k\right)$ via hidden DNN layers at $t$-th epoch to evaluate the selected action, which is written as
\begin{align}\label{Q function}
	Q\left(s_{k,t},a_{k,t}|\theta^{Q}_k\right)=\mathbb{E}\left[r_k+\gamma Q\left(s_{k,t+1},a_{k,t+1}|\theta^{Q}_k\right)\right],
\end{align}		
where $\gamma$ is a discount factor and $\mathbb{E}\left[\cdot\right]$ is the expectation for the trajectory since it is difficult to sample thousands of situations. In order to update the DDPG network, the gradient of actor network is acquired as
\begin{equation}	
\begin{aligned}\label{actor function}	   						\nabla_{\theta^{\mu}_k}J_i=&\nabla_{\theta^{\mu}_k}\mathbb{E}\left[Q\left(s_{k,t},a_{k,t}|\theta^{Q}_k\right)\right]\\
	=&\mathbb{E}\left[\nabla_{a_{k,t}}Q\left(s_{k,t},a_{k,t}|\theta^{Q}_k\right)\cdot\nabla_{\theta^{\mu}_k}\mu\left(s_{k,t}|\theta^{\mu}_k\right)\right],
\end{aligned}		
\end{equation}
where critic loss function can be given by
\begin{align}\label{critic loss function}
	L_k&=\mathbb{E}\left[y_k-Q\left(s_k,a_k|\theta^{Q}_k\right)\right]^{2} 
\end{align}	
with $y_k=r_k+\gamma Q\left(s_{k,t+1},a_{k,t+1}|\theta^{Q'}_k\right)$. The target network will periodically update the network weights from the main network based on the soft update \cite{LTPHJPPAH15} for both actor-critic sub-networks which is represented by
\begin{align}\label{soft network}	\theta^{Q'}_k&=\tau_{a}\theta^{Q}_k+(1-\tau_{a})\theta^{Q'}_k,\\
\theta^{\mu'}_k&=\tau_{c}\theta^{\mu}_k+(1-\tau_{c})\theta^{\mu'}_k,
\end{align}	
where $\tau_{a}$ and $\tau_{c}$ are constants indicating the significance of parameters in the target and main networks.


\begin{algorithm}[!t]

\caption{Proposed FDRL Algorithm}
\SetAlgoLined
\DontPrintSemicolon
\label{alg}
	\begin{algorithmic}[1]
		\STATE Input: $\mathbf{h}_{kk}^{lim}$, $\Gamma_k, \forall k$
		\STATE Output: $a_k=\mathbf{w}_k, \forall i$
		\STATE Initialize: $\theta_k^{\mu}$, $\theta_k^{\mu'}$,  
$\theta_k^{Q}$, $\theta_k^{Q'}$ $\forall k$, $\theta^{global}$, replay memory $M_r$
		\FOR{$t=1,2...,E$}
		\FOR{each BS $k$}
			\STATE{Decide the action $a_{k,t}=\mu(s_{k,t}|\theta^{\mu}_k)+N_{G}$}
			
			\STATE{Interact with the environment and save result of $\left(s_{k,t},a_{k,t},r_k,s_{k,t+1}\right)$ to replay memory $M_r$}
			\STATE{Off-line actor/critic model training by \\mini-batching data with a size of $B$} 
			\STATE{Soft update $\theta_k^{\mu'}, \theta_k^{Q'}$} 
		\ENDFOR	
		\IF{$\mod (t, T)=0$}
			\STATE{FEL model aggregation:  $\theta^{global}=\frac{1}{K}\sum_{i=1}^{K}\xi_k\theta_k^{local}$}
			\STATE{Model update after aggregation: $\theta_k^{local}=\theta^{global}$}
		\ENDIF
		\ENDFOR
	\end{algorithmic}
	
\end{algorithm}


\subsection{Federated Edge Learning for Interference Mitigation}

After DDPG learning for THz beamforming adjustment, the local DDPG training models of NN’s weights will be sent from
BSs to the edge server per $T$ iteration. The edge server will then aggregates local training model in order to exchange the hidden information of interference information rather than directly upload high-overhead full CSI data. Inspired by the FEL method in \cite{LTPHJPPAH15}, the model aggregation can be presented as
\begin{align}\label{weighted fed}
	\theta^{global}=\frac{1}{K}\sum_{k=1}^{K}\xi_k\theta_k^{local},
\end{align}
where $\xi_k\in[0,1]$ is a ratio indicating the importance of each training model depending on certain property of dataset in each BS and $\sum_{k=1}^{K}\xi_k=1$. In $\eqref{weighted fed}$, $\theta_k^{local}=\left\lbrace \theta^{\mu}_k, \theta^{Q}_k\right\rbrace$ consists of neural weights of  main actor/critic network. Note that in this case, we consider $\xi_k=1$ as equivalent importance of each beamforming model since the THz BS could provide potentially useful information of limited estimated THz CSI. After finishing the model aggregation, the edge server returns the global parameters to each BS, which is repeatedly performed until convergence. Thus, the candidate beamforming of each THz BS $\mathbf{w}_k$ will converge to near optimum by searching for the higher DDPG reward through the iterative training. Additionally, to further
address full model upload problem, we design a partial FEL mehcanism by uploading parital weight elements $\theta_{k,P}^{local}$ of original training model of $\theta_k^{local}$. That is, we upload the weight elements with a ratio of $\mathcal{N}(\theta_{i,P}^{local}) / \mathcal{N}(\theta_{i}^{local})$, where $\mathcal{N}(\cdot)$ is defined as a function calculating the number of elements in the weight vectors. The concrete algorithm of proposed FDRL is proposed in Algorithm \ref{alg}.

\section{Simulation Results} \label{SR}

In this section, we have performed simulations of proposed FDRL-enabled THz beamforming with maximization of system throughput while mitigating network interference. The THz BS and serving UEs are uniformly-randomly distributed in the radius from 10 to 100 meters. We consider $N_{NL}=5$ NLoS paths and THz frequency is set to be $0.3$ THz. The remaining parameter setting of the THz network is listed in Table \ref{Parameter}.

\begin{figure}
	\centering
	\includegraphics[width=3.2in, height=2.2in]{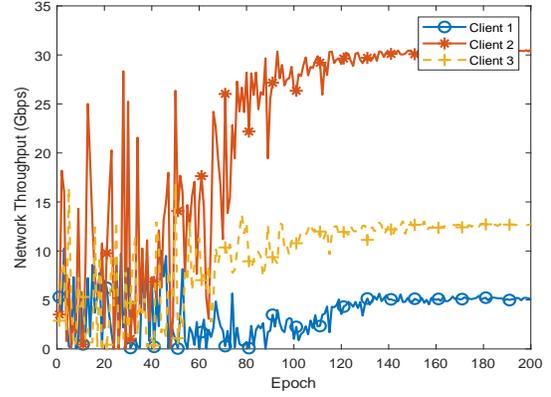}
	
	\caption{The network throughput achieved by the proposed FDRL with 3 BSs (i.e., $K=3$) along the training epochs.} \label{Fig.3}

\end{figure}

In Fig. \ref{Fig.3}, the convergence of the throughput with $K=3$, clients served by three BS equipped with $N_t=8$ antennas. Each BS have a two-layer actor-critic neural network with $\left\lbrace100,70 \right\rbrace$ neurons. The actor network decides the beamforming vector $\mathbf{w}_k$, while the critic network using the Q-learning network evaluates the decision of actor network. Initially, the beamforming vector is randomly selected with the perturbation $N_G$ with variance equal to $3$ leveraging exploitation and exploration. However, it will gradually decay to $0.99$ as deterministic decision. At around $100$-th epoch, the client $1$ tends to be stable, but the others are still searching for the potential solution from DDPG network. At around $150$-th epoch, the performance of throughput is almost converged. Note that the result only shows a single run of certain channel condition in Fig. \ref{Fig.3}; however, we will conduct 100 Monte Carlo runs leveraging different channel conditions in the following comparisons.
	

\begin{table}
    \fontsize{10}{10}\selectfont
    \vspace{-2.0em}
	\begin{center}
		\caption {Parameters of FDRL-enabled THz Network }
		
		\resizebox{\columnwidth}{!}{\begin{tabular}{|l|c|l|}
				\hline
				
				Definition & Symbol & Value \\ \hline 
				
				Discount factor & $\gamma$ & $0.9$ \\
				\hline
				Significance of actor network & $\tau_{a}$ & $0.01$\\
				\hline        
				Significance of critic network & $\tau_{c}$ & $0.01$\\
				\hline 
				Number of epoch & $E$ & $300$\\
				\hline 
				
				Buffer of the memory size & $M_r$ & $10$\\
				\hline 
				Batch size of DDPG training & $B$ & $5$\\
				\hline 
				The cycle of FEL & $T$ & $20$\\
				\hline
				Number of antennas & $N$ & $\{8,16,32,64,128,256\}$\\
				\hline
				Number of BSs & $K$ & $\{2,3,6\}$\\
				\hline
				Operating frequency & $f$ & $0.3$ THz\\
				\hline
				Distance between BS and UE & $d$ & $[10,100]$ m\\
				\hline
				Medium absorption factor & $\rho(f)$ & $0.1$\\
				\hline	
				Number of NLoS paths & $L$ & $5$ \\
				\hline	
				Integrated antenna gains & $G$ & $10$ dB\\
				\hline
				Noise power & $\sigma^2_n$ & $-74$ dBm\\
				\hline	
				Transmit power & $P$ & $10$ dBm\\
				\hline
		\end{tabular}} \label{Parameter}
	\end{center}
	\vspace{-0.2in}
\end{table}

\begin{figure}[!t]

\subfigure[Complexity of FDRL]{
\begin{minipage}[t]{0.2\textwidth}
\centering
\includegraphics[width=4.4cm]{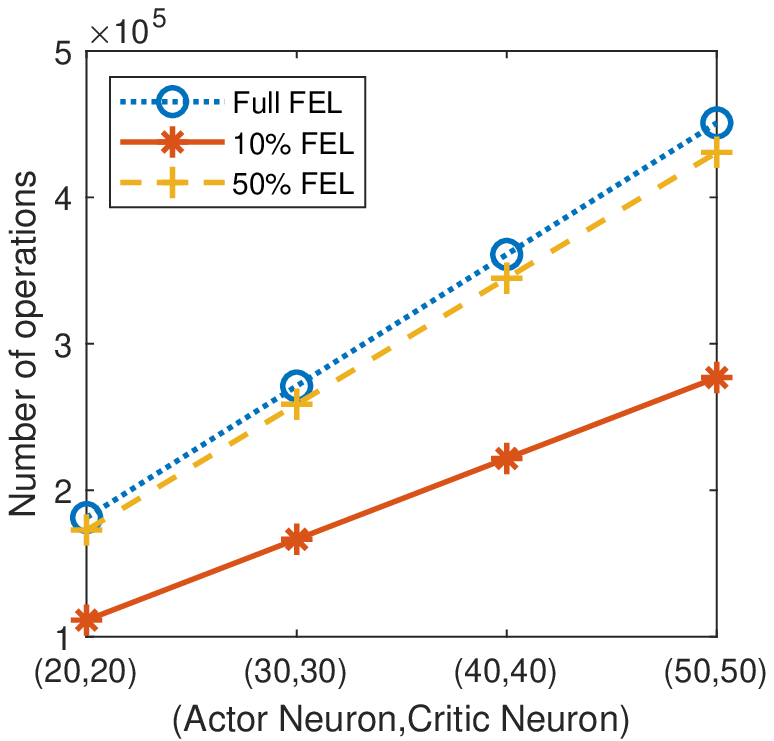}

\end{minipage}
}
\subfigure[Throughput of FDRL]{
\begin{minipage}[t]{0.2\textwidth}
\centering
\includegraphics[width=4.4cm]{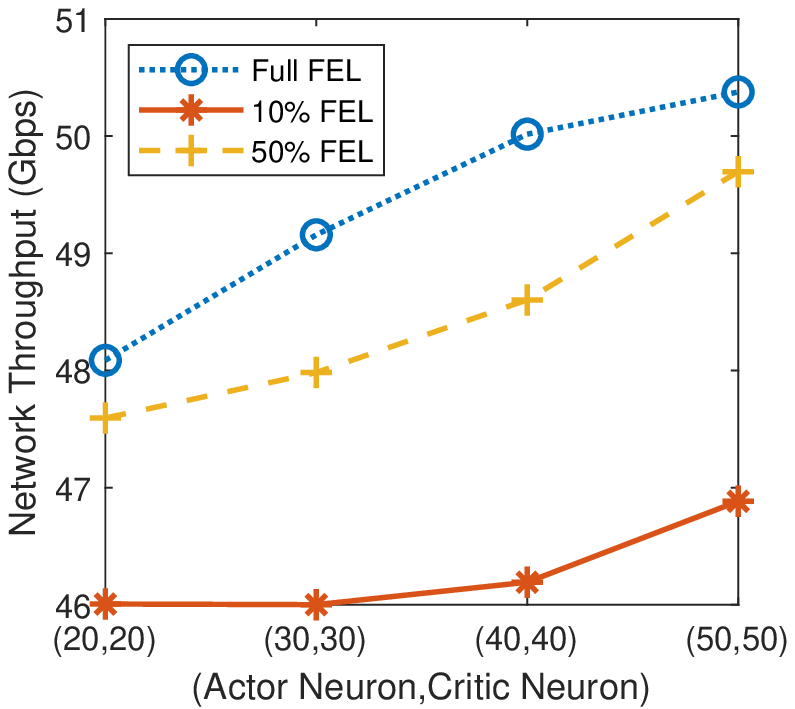}

\end{minipage}
}
\vspace{-0.15in}
\caption{The proposed FDRL compared to different numbers of actor and critic neurons with $\left\lbrace (20,20),(30,30),(40,40),(50,50) \right\rbrace$.}
\label{Fig.4}

\end{figure}

In Fig. \ref{Fig.4}, we illustrate the throughput that is affected by the number of the actor and critic neurons. Each of the BSs is equipped with $N_t=128$ antennas and $K=4$ is considered. As the actor-critic neurons is set as $(20,20)$, the performance of full FEL upload achieves higher throughput than that of partial upload with around $2$ and $0.5$ Gbps for $10\%$ and $50\%$ upload of FEL parameters. However, the operational overhead, i.e., operation for training in local network and FEL server is quite higher using full upload than that of $10\%$ upload. We can observe that the $10\%$-upload overhead possesses half the overhead compared to that of full FEL upload while sustaining adequately high throughput performance, i.e., the $10\%$ exchanged hidden information is enough to alleviate induced interference. Moreover, when the number of actor-critic neurons is increased to $(30,30)$, it reaches high throughput performance but provokes higher computational overhead,  which strikes a compelling tradeoff between complexity and throughput performance.

\begin{figure}[!t]
\centering
\includegraphics[width=3.2in, height=2.2in]{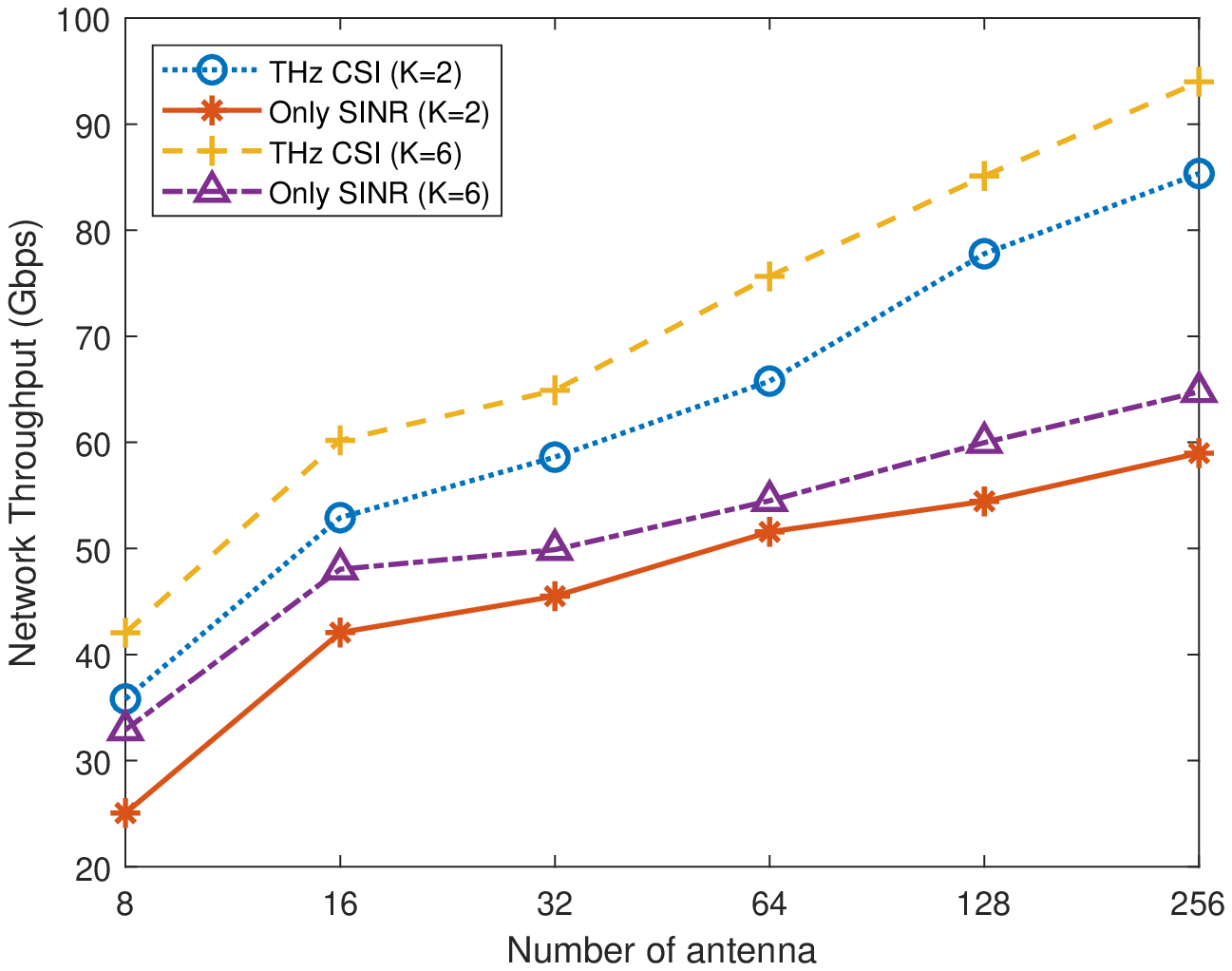}

\caption{Network throughput of the proposed FDRL with different number of antennas considering limited CSI and only SINR feedback with $K =\left\lbrace2,6\right\rbrace$ BS.} \label{Fig.5}
\vspace{-0.3in}
\end{figure}

Fig. \ref{Fig.5} demonstrates the throughput considering input states of CSI and only SINR feedback with different number of antennas and THz BS deployment  with $K=\{2,6\}$ and  $N_t=\{8,16,32,64,128,256\}$ antennas. We can observe the result of $K\!=\!2$ and $K\!=\!6$ that higher performance can be obtained due to advantageous FDRL of interference mitigation. Furthermore, the throughput performance is proportional to the number of antennas due to higher spatial diversity. In addition, throughput difference becomes increasingly larger when $N_t=256$ antennas are equipped by comparing the mechanism of THz CSI and of only SINR feedback. This is because that more hidden information from the estimated CSI is extracted and exchanged by FEL server, which is advantageous to interference cancellation. 

\begin{figure}
\centering
\includegraphics[width=3.2in, height=2.2in]{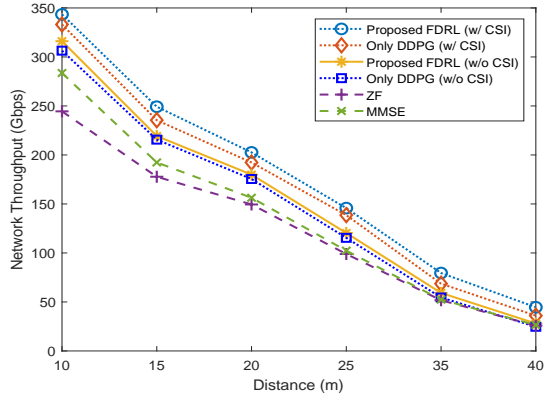}

\caption{The proposed FDRL algorithm compared to conventional and non-FEL benchmark optimization methods.} \label{Fig.6}

\end{figure}

Fig. \ref{Fig.6} demonstrates the throughput considering distances between THz BS-UE. The throughput of all algorithms decreases to near zero due to limitation of severe intrinsic path loss from the THz channel. The proposed FDRL algorithm with estimated CSI can exchange more hidden information than that without CSI, which achieves higher throughput performance. In addition, DDPG lacks training model exchange from FEL because each BS conducts local training without model exchange, which results in lower throughput than the proposed FDRL algorithm. Moreover, the proposed FDRL is capable of exchanging sufficient hidden training models from powerful deep learning based DDPG, which outperforms the conventional beamforming methods of zero forcing (ZF) and minimum mean square error (MMSE).

\section{Conclusions} \label{CON}

In this paper, an FDRL scheme was proposed to improve the network throughput by optimally searching the beamformers of the BSs under the situation of limited THz CSI. We have numerically demonstrated that FDRL is capable of exchanging more representative features among THz BSs to alleviate interference as well as to achieve higher throughput even when the BSs have limited CSI. With more deployed antenna arrays, the network is able to achieve a higher throughput due to higher spatial diversity. Moreover, we can observe a compelling tradeoff between overhead of exchange information and network throughput. Namely, the network can achieve high throughput at a low cost of uploading partial FDRL models. In a nutshell, the proposed FDRL scheme using DDPG and FEL architectures outperforms the baseline methods with pure deep Q-learning because it can take the advantage of interference mitigation from information exchange. Also, we show that the proposed FDRL is superior to the existing beamforming techniques using non-learning methods, such as ZF and MMSE.

\bibliographystyle{IEEEtran}
\bibliography{IEEEabrv,Ref_FDRL_THz}
\end{document}